\documentclass[a4paper, 10pt, conference]{IEEEtran}
\IEEEoverridecommandlockouts
\usepackage{cite}
\usepackage{amsmath,amssymb,amsfonts}
\usepackage{graphicx}
\usepackage{textcomp}
\usepackage{algpseudocode}
\usepackage{algorithm}
\usepackage{hyperref}
\usepackage{float}
\usepackage{array}
\usepackage{color,colortbl}
\definecolor{Gray3}{gray}{0.9}
\definecolor{Gray5}{gray}{0.1}
\definecolor{Gray2}{gray}{0.7}
\definecolor{Gray1}{gray}{0.5}

\usepackage{xcolor}
\def\BibTeX{{\rm B\kern-.05em{\sc i\kern-.025em b}\kern-.08em
    T\kern-.1667em\lower.7ex\hbox{E}\kern-.125emX}}
\begin{document}

\title{SwarMan: Anthropomorphic Swarm of Drones Avatar with Body Tracking and Deep Learning-Based Gesture Recognition\\
\thanks{The reported study was funded by RFBR and CNRS according to the research project No. 21-58-15006.}
}
\makeatletter
\newcommand{\linebreakand}{%
  \end{@IEEEauthorhalign}
  \hfill\mbox{}\par
  \mbox{}\hfill\begin{@IEEEauthorhalign}
}
\makeatother
\author{\IEEEauthorblockN{Ahmed Baza}
\IEEEauthorblockA{\textit{Digital Engineering Center} \\
\textit{Skoltech}\\
Moscow, Russia \\
ahmed.baza@skoltech.ru}
\and
\IEEEauthorblockN{Ayush Gupta}
\IEEEauthorblockA{\textit{Digital Engineering Center} \\
\textit{Skoltech}\\
Moscow, Russia \\
ayush.gupta@skoltech.ru}
\and
\IEEEauthorblockN{Ekaterina Dorzhieva}
\IEEEauthorblockA{\textit{Digital Engineering Center} \\
\textit{Skoltech}\\
Moscow, Russia \\
ekaterina.dorzhieva@skoltech.ru}

\linebreakand
\IEEEauthorblockN{Aleksey Fedoseev}
\IEEEauthorblockA{\textit{Digital Engineering Center} \\
\textit{Skoltech}\\
Moscow, Russia \\
aleksey.fedoseev@skoltech.ru}
\and
\IEEEauthorblockN{Dzmitry Tsetserukou}
\IEEEauthorblockA{\textit{Digital Engineering Center} \\
\textit{Skoltech}\\
Moscow, Russia \\
d.tsetserukou@skoltech.ru}
}

\maketitle

\begin{abstract}
Anthropomorphic robot avatars present a conceptually novel approach to remote affective communication, allowing people across the world a wider specter of emotional and social exchanges over traditional 2D and 3D image data. However, there are several limitations of current telepresence robots, such as the high weight, complexity of the system that prevents its fast deployment, and the limited workspace of the avatars mounted on either static or wheeled mobile platforms. 

In this paper, we present a novel concept of telecommunication through a robot avatar based on an anthropomorphic swarm of drones; SwarMan. The developed system consists of nine nanocopters controlled remotely by the operator through a gesture recognition interface. SwarMan allows operators to communicate by directly following their motions and by recognizing one of the prerecorded emotional patterns, thus rendering the captured emotion as illumination on the drones. 
The LSTM MediaPipe network was trained on a collected dataset of 600 short videos with five emotional gestures. The accuracy of achieved emotion recognition was 97\% on the test dataset.

As communication through the swarm avatar significantly changes the visual appearance of the operator, we investigated the ability of the users to recognize and respond to emotions performed by the swarm of drones. The experimental results revealed a high consistency between the users in rating emotions. Additionally, users indicated low physical demand (2.25 on the Likert scale) and were satisfied with their performance (1.38 on the Likert scale) when communicating by the SwarMan interface.  

\end{abstract}


\section{Introduction}
% Dorzhieva

With the latest development in robotics and telepresence technology, along with the production of robots for manufacturing purposes, more attention is paid to the use of robots in everyday life. 
\begin{figure}[!h]
\centering
\includegraphics[width=0.9\linewidth]{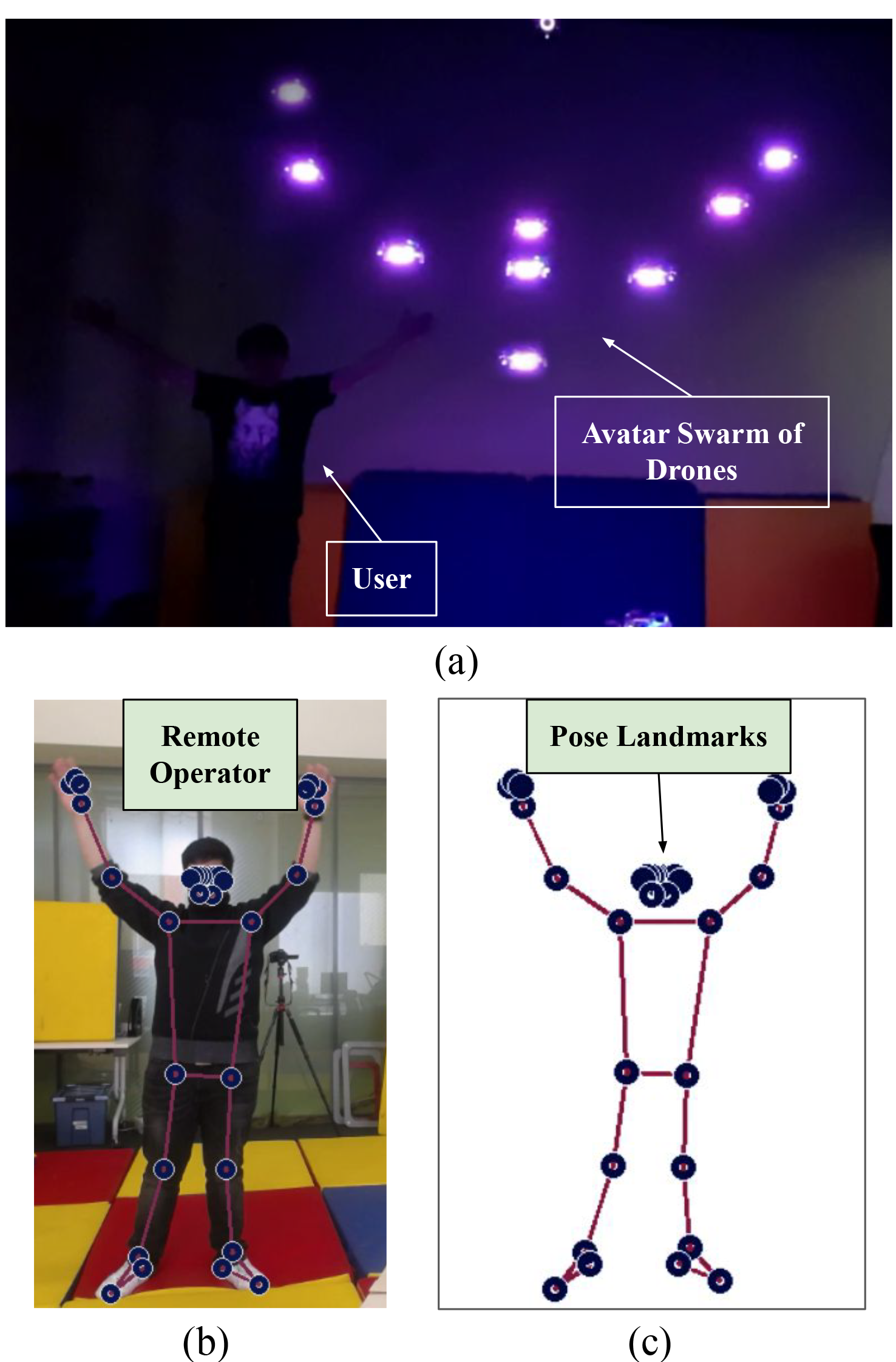}
\caption{(a) User interaction with SwarMan avatar. (b) The remote avatar performs gestures in front of the camera. (c) Point landmarks of the recognized "Happy" gesture.} \label{system}
\end{figure} 
Novel research topics are emerging in the field of service robots, suggesting their application as companions to improve the mental state of humans. To achieve the necessary behavior complexity, robots have to accurately determine the state of the user to establish natural communication.

For example, Muhammad Abdullah et al. \cite{9501446} developed the emotion recognition system that uses the voice features in addition to the facial expressions of a human for the robot assistant functionality.

Companion robots can play with children and teach them, as proposed in the research by Leite et al. \cite{6249581} in which the developed robots responded empathetically to several of the children's affective states. 
In addition to the voiced indication of certain emotions and the body language, several papers are focused on robots that can broadcast the emotional state by their eyes, e.g., an eyeball robot developed by Shimizu et al. \cite{8371114}.

% transfer emotions from operator to user via robots, telepresence 
% However, the emotional communication between the robot and the human has encountered difficulties, such as a lack of conceptual understanding. The robot still cannot respond like a human but are competent to broadcast the emotions of the operator to another user. The task of telecommunication is widely considered in the application of virtual and augmented reality. Soomro et al. proposed wearable displays \cite{8446471}, while Lee et al. \cite{7892320} developed an augmented telepresence platform to exchange sensory information. But in various situations, when, for example, physical contact with the user or surrounding objects is required, virtual reality is not valuable.

% about agents
% Mobile robots can be used as agents in teleoperation and telepresence tasks, but it is plausible that most of these robots are designed to resemble the human body and perform various similar operations \cite{hanson2020neurosymbolic}. However, telecommunication through the robotic avatar requires delivering the robot to the working area, which is often proves challenging either due to bulkiness of the robot or due the dangerous environment. Swarm of drones are promising remote control tools, as in SwarmPaint \cite{serpiva2021swarmpaint} that uses a swarm of gesture-controlled drones to paint in the air.

Mobile robots are actively used as agents in teleoperation and telepresence tasks for affective communication. Most of these robots are designed to resemble the human body and to perform various operations similar to a human \cite{hanson2020neurosymbolic}. However, telecommunication through the robotic avatar requires delivering the robot to the working area, which often proves challenging either due to the bulkiness of the robot or due to the dangerous environment. The operator stations have been equipped to organize the work of stationary robots, as suggested in the research of Christian Lenz and Sven Behnke \cite{lenz2021bimanual}, for telemanipulation by anthropomorphic avatar arms.

The mentioned above scenarios propose highly sufficient robotic systems. However, the mobility of the mentioned above robots is strictly limited by the workspace of the robot's upper body and the physical dimensions of the mobile platform. Moreover, their implementation may be challenging to the user due to the high mass and relatively slow operation of these systems. Meanwhile, a swarm of drones can serve as an effective remote-control tool. Several researchers explored applications of the robotic swarms in teleoperation, for example, Serpiva et al. \cite{serpiva2021swarmpaint} with the SwarmPaint system that utilizes a swarm of gesture-controlled drones to change formations and paint by the light in the air. Recently, due to the fast developments in telepresence technologies alongside virtual and mixed reality technologies, the teleoperation of drones avatars is suggested by Cordar et al. \cite{virtual_humans} for human telepresence and foster empathy with virtual agents and robots

In this paper, we propose a novel approach to the task of telepresence, involving a swarm of drones in broadcasting emotions from the operator to the user.

\section{Related Works} 
%Aleksey
Anthropomorphic robot avatars were extensively investigated and improved in recent years. Such systems as TELESAR VI developed by Tachi et al. \cite{Tachi_2020} allow dexterous remote manipulation and communication through an avatar designed to resemble the upper body of the human. 

Several researchers investigated effective communication through robot avatars. For example, Tsetserukou et al. \cite{Tsetserukou_2010} explored remote affective communications and proposed the robotic haptic device iFeel IM to augment the emotional experience during online conversations. Bartneck et al. \cite{Bartneck_2004} explored the dependence of human emotion perception on the character’s embodiment, showing that there is no significant difference in the perceived intensity and recognition accuracy between robotic and screen characters. Chao-gang et al.\cite{Chao-gang_2008} proposed a facial emotion generation model based on random graphs for virtual robots. A fuzzy emotion system that controls the face and the voice modules was developed by Vasquez et al. \cite{Vasquez_2020} for a tour-guide mobile robot.

Though facial expressions play a major role in emotional recognition, the dynamic body postures could be recognized with relatively high precision. Matsui et al. \cite{Matsui_2005} proposed a motion mapping approach to generate natural behavior for humanoid robots by copying human gestures. Cohen et al. \cite{Cohen_2011} explored children's reactions to the iCat and NAO robots and achieved to design of well-recognized body postures for NAO. The end-to-end neural network model was developed by Yoon et al. \cite{Yoon_2019} to generate sequences of human-like gestures enhancing NAO speech content. The variational auto-encoder framework was implemented by Marmpena et al. \cite{Marmpena_2019} for generating numerous emotional body language for the anthropomorphic Pepper robot.

\section{System Architecture}

% Ayush Gupta 

In the developed architecture shown in Fig. \ref{system} the user interacts with the avatar swarm by visual interpretation of the emotion while the avatar operator performs the various body pose gestures to operate the avatar swarm of drones. The tracking and localization of the drones are done through the VICON mocap system which consists of 12 infra-red (IR) cameras. 

\begin{figure}[htp]
\centering
\includegraphics[width=1\linewidth]{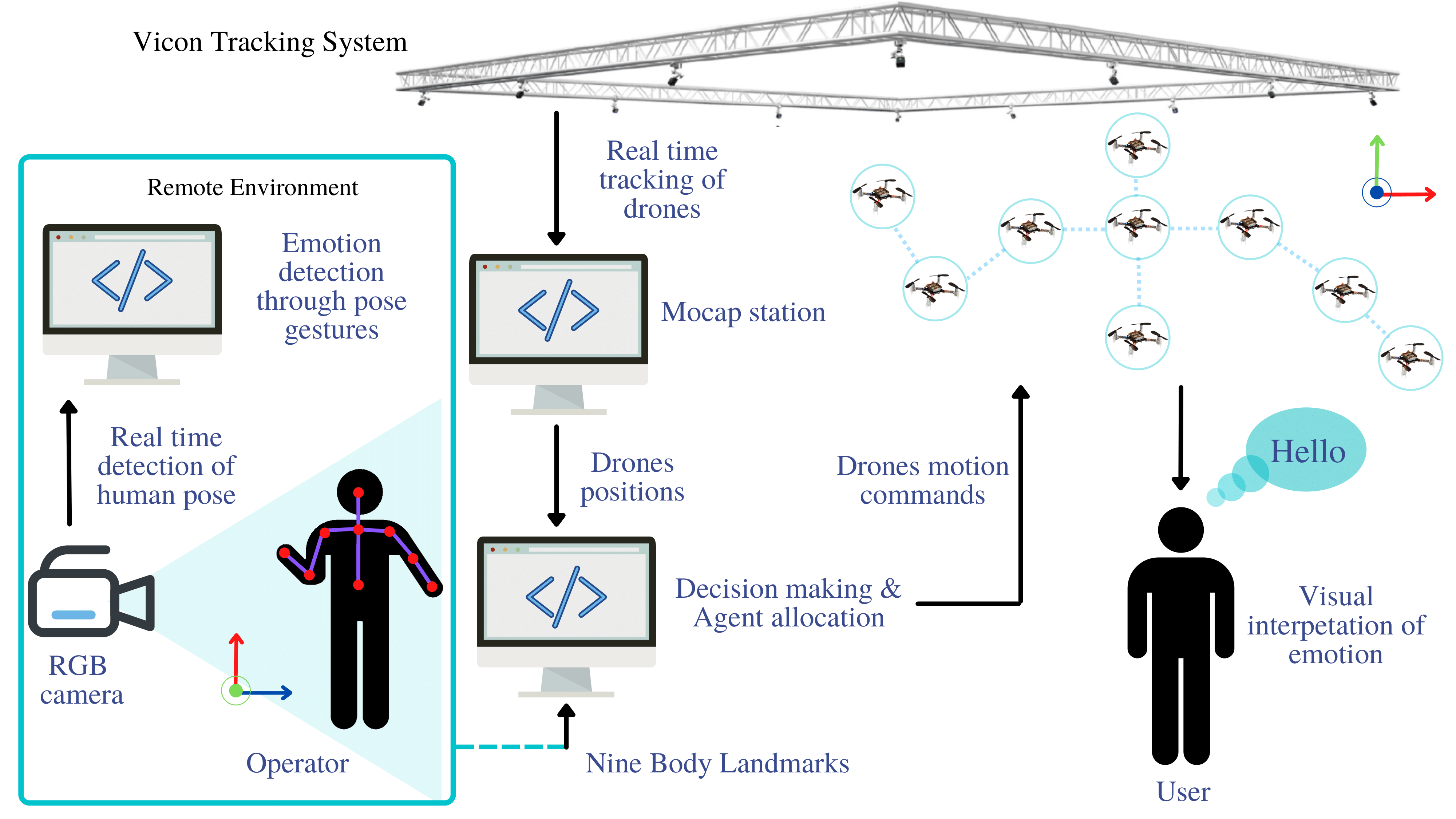}
\caption{Layout of the SwarMan system.} \label{system}
\end{figure}

In the remote environment, the operator performs various gestures which showcase different emotions. These poses are captured by a DL-based gesture recognition algorithm which includes the major upper-body nine landmarks which include head, neck, left-shoulder, right-shoulder, left-elbow, right-elbow, right-hand and left-hand. These landmarks are then passed to the decision-making and agent allocation algorithm where along with the localized positions of the swarm the designated positions of the swarm of drones are calculated according to the relative positions of the major nine joints of the human upper body pose. The user interacts with the swarm of drones visually to understand the emotions that the operator was trying to perform. Along with the different poses of the emotions, the light rings on the drones also convey a psychological effect on the user for interpreting the type of emotion which includes green for happy, red for angry, white for neutral, yellow for confusion, and blue for sad.

\section{Trajectory Generation and Swarm Control} \label{trajectory}
% Ahmed Baza
For a more immersive experience and intuitive control, the operator of the avatar is controlling the swarm of drones through a camera feed. The operator's body landmarks are collected and then used for trajectory generation and Gesture recognition (\ref{gesture}). Based on the calculated trajectory, each drone is assigned a role to fly as in the swarm, e.g., left hand, right shoulder, and head. 

\subsection{Trajectory Generation}
The operator's body is being tracked using the MediaPipe Holistic pipeline and the resulting body landmarks are extracted from the camera feed. Since the body landmarks are in the camera frame, both the landmarks' positions and the body scale are dependent on the operator's distance from the camera. Transformation of the axis is needed to represent the swarm in the real remote environment. Let $S$ be the camera coordinate frame; we assume a coordinate frame $S'$ with an origin at the head landmark in the camera frame; thus the position $p'$ of a body landmark $p$ relative to $S'$ is given by:

\vspace{-0.4em}
\begin{equation}
 \label{eq:1}
 p' = p - h
\end{equation}
where $h$ is the head position relative to $S$, $p$ is the body landmark position. 
Using the resulting points, a tree of vectors is constructed, shown in (Fig. \ref{fig:vec}).

\begin{figure}[!h]
 \centering
 \includegraphics[width=0.8\linewidth]{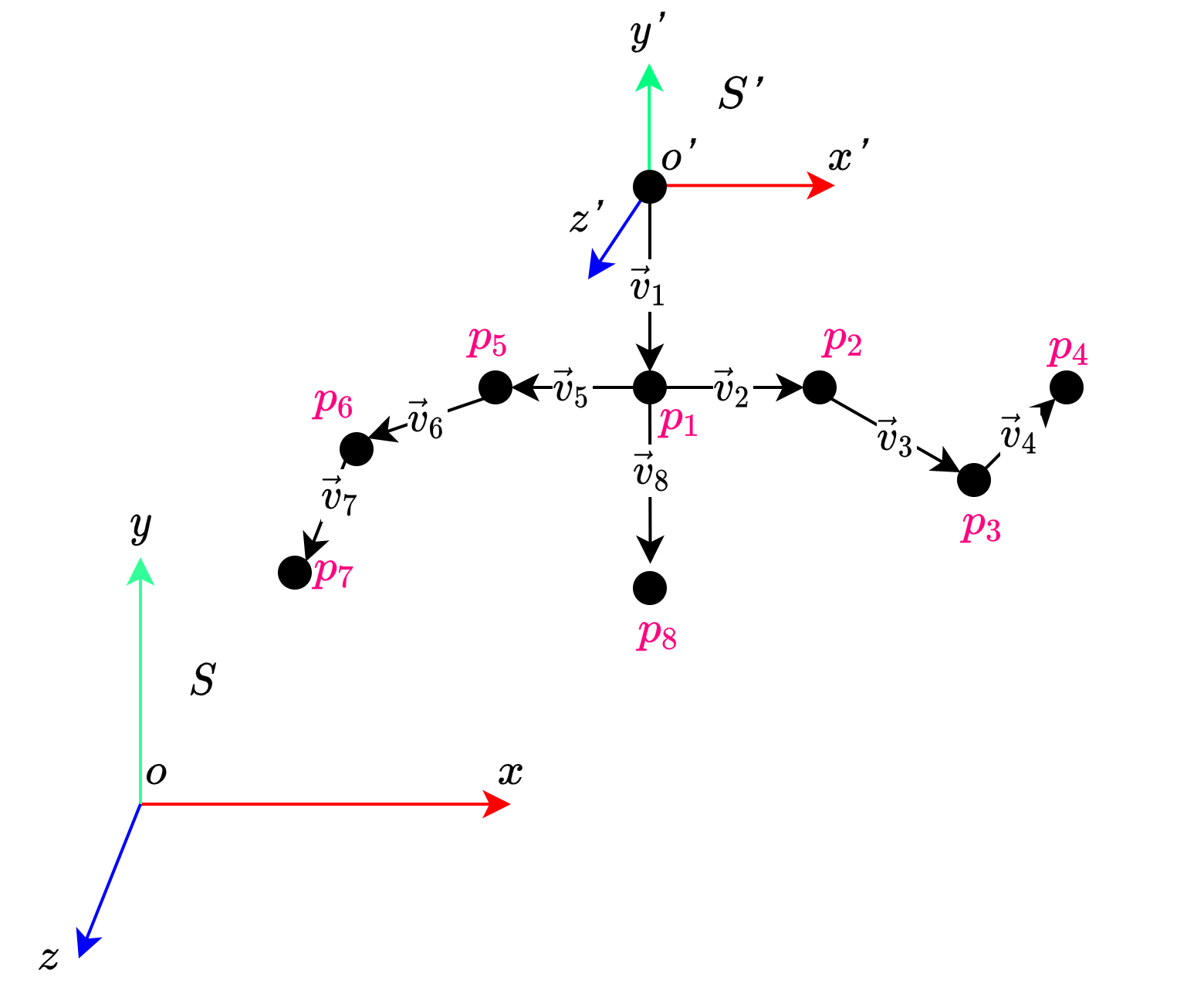} % nice 
 \caption{Vector representation of the tracked body landmarks with coordinate frames.}
 \label{fig:vec}
\end{figure}

The vectors $\left[\overrightarrow{v_{1}},\overrightarrow{v_{2}}, ...,\overrightarrow{v_{8}}\right]$ are the unit vectors pointing from each parent landmark to child landmark, relative to $S'$, and is described as:

\begin{equation}
 \label{eq:2}
\overrightarrow{v_{m}} =\frac{p'_{m} - p'_{n}}{\parallel p'_{m} - p'_{n}\parallel} ,
\end{equation}
where $p'_{n}$ is the parent landmark position, $p'_{m}$ is the child landmark position.

An accurately body-scaled formation can be obtained by multiplying each unit vector by the actual length of the correspondent operator's body part, thus eliminating the change in swarm size with the change in the operator-camera distance. 
Since the list $\left[\overrightarrow{v_{1}},\overrightarrow{v_{2}},...,\overrightarrow{v_{8}}\right]$ represents the positions of the target formation in the frame $S'$, where the head landmark is the origin, then by choosing the position of head drone in the real environment we can represent each target in the formation using the recursive equation shown in: 

\begin{equation}
 \label{eq:3}
P_{m} =\overrightarrow{v_{m}} * L_{m}+P_{n}
\end{equation}
where $P_{m}$, $P_{n}$ and $L_{m}$ are the child point target in the real environment, the parent point target in the real environment, and the length of the correspondent operator's body part, respectively. $P_{0}$ is equal to the needed head position in the real environment.

\subsection{Agent Assignment}
Crazyflie drones are used in the system utilizing the Crazyswarm ROS package, the package offers some reliable solutions for swarm management. integrating with Crazyswarm solutions, dynamic agent assignment helps the system be less complex, easy to set up, and more reliable. The drones are starting at random positions and are assigned to target positions in the formation based on the minimum value of a cost function. Euclidean distance from the drone's starting position to the target position is used as a cost function. The protocol is explained in algorithm \ref{alg:1}.

\begin{algorithm}
\caption{Agent Assignment Algorithm}\label{alg:1}
\begin{algorithmic}[1]
\State $Initiate$
\State $Swarm = [Drone_1, Drone_2, ..., Drone_9 ]$ 
\State $Targets = [P_1, P_2, ..., P_9]$ 
\State $Assigned = []$
\For {$P$ in $Targets$}
   \State $costs = []$ 
    \For{$Drone$ in $Swarm $ }
        \If {$Drone$ NOT in $Assigned$}
            \State $cost = f(P,Drone)$
            \State $costs.append(cost)$
        \Else 
            \State $Pass$
        \EndIf
    \EndFor
    \State $i = $index$ $of$(min(costs))$
    \State $Assigned.append(Swarm[i])$
    \State $Swarm[i].target = P$
\EndFor
\end{algorithmic}
\end{algorithm}

The Euclidean distance $f$ is implemented as a cost function, which is calculated as given by:

\begin{equation}
 \label{eq:4}
f(p, q) =\sqrt{\sum_{i=1}^3(q_{i}-p_{i})^2} 
\end{equation}
where $p_{i}$ is the target landmark position, $q_{i}$ is the position of the drone of the swarm. The result of the task assignment is shown in Fig. \ref{assignment}.

\begin{figure}[htp]
\centering
\includegraphics[width=1\linewidth]{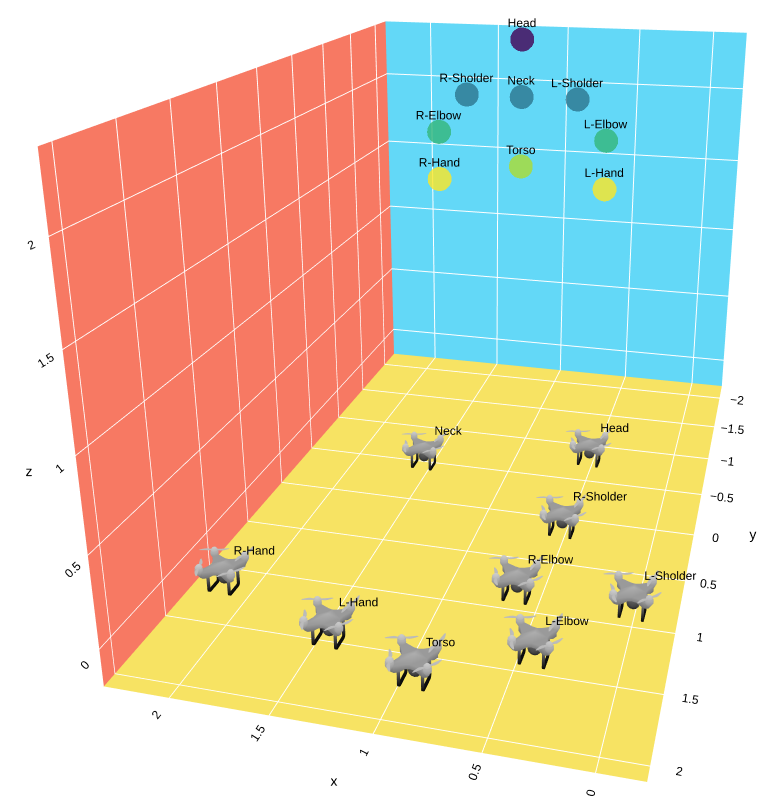}
\caption{Agent assignment results: Drones in random starting position assigned to target formation.} \label{assignment}
\end{figure}

The successful agent assignment is achieved with the developed algorithm by finding the nearest available drone to the formation target.

\subsection{Collision Avoidance}
To avoid the inter-agent collision, Artificial Potential Field (APF) was implemented. The APF algorithm is simply modeling the target way-point as an attraction field and obstacles, other swarm agents, as repulsion filed. A navigation policy is constructed by calculating the resultant virtual force on the agent, using it to steer the drone away from obstacles toward the target. Since the drones are flying in a close formation, the repulsive potential is modeled to only act within a sphere of influence surrounding each drone. The radius $r_{0}$ of the implemented sphere of influence, \ref{eq:7}, where $[0.2, 0.2, 0.4]^T$ as the down-wash of drones were affecting each other, the $z$ component was increased compared to the $x$ and $y$ components.
The overall potential $U_{sum}$, shown in:  
\vspace{-0.4em}
\begin{equation}
 \label{eq:5}
 U_\text{sum}(x,y,z) =U_\text{att}(x,y,z) +U_\text{rep}(x,y,z)
\end{equation}
%\vspace{0.5em}
where $U_{att}$ is the attraction potential and $U_{rep}$ is the repulsion potential. The attraction potential is defined in:

\begin{equation}
 \label{eq:6}
U_{att}(x,y,z)=\xi \left \| P_{drone}-P_{target} \right \|^{2}
\end{equation}
where $ P_{drone}$ is the drone current position, $P_{target}$ is the drone desired position, and $\xi$ is the scaling factor. The repulsive potential is defined in: 

\begin{equation}
 \label{eq:7}
 U_\text{rep}(x,y,z) = \begin{cases}
 \frac{1}{2}\eta(\frac{1}{\rho(x,y,z)} - \frac{1}{r_\text{0}})^2 \quad &, \hspace{2mm}\rho \leq r_\text{0} \\
 0 \quad & ,\hspace{2mm}\rho > r_\text{0} \\
 \end{cases}
\end{equation}
where $\rho(x,y,z)$ is the distance function and $\eta$ is the constant scaling factor. The distance function $\rho(x,y,z)$ is implemented as Manhattan distance described as:

\begin{equation}
 \label{eq:8}
\rho(p, q) =\sum_{i=1}^3\mid q_{i}-p_{i} \mid
\end{equation}
where $p$ and $q$ are the drone position and the obstacle position, respectively. 
% -------------------------------------------------------------------
%                       END OF SECTION
% -------------------------------------------------------------------
\section{Gesture Recognition}  \label{gesture}
% Ekaterina Dorzhieva

In the process of interaction between a swarm of drones and a user, we used five simple, most commonly expressed emotions during communication: happiness, sadness, anger, confusion, and neutrality shown in Fig. \ref{emotions}. To achieve an immersive experience for the operator as well as the user gesture recognition is added based on body tracking. While the user communicates through body tracking, rendering his motion into avatar poses representing his feelings, the system can help the operator convey the mentioned basic emotions through drone illumination utilizing gesture recognition. Thus, when the operator wants to convey to the user happiness, a simple victory hands up is recognized by our model while the swarm of drones repeats the operator's movements, it changes the color of the agents' light ring to green as a hint to the user. The system renders white, blue, red, and yellow to convey neutrality, sadness, anger, and confusion, respectively.

\begin{figure*}[htp]
\centering
\includegraphics[width=1\linewidth]{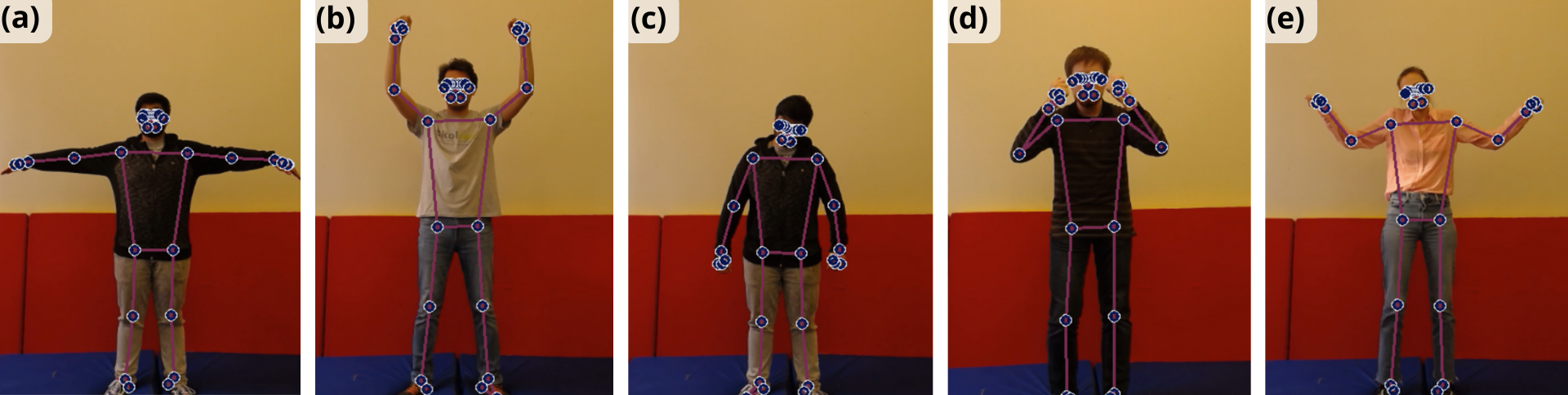}
\caption{Example of gestures representing emotions. a) Neutral. b) Happy. c) Sad. d) Angry. e) Confused.} \label{emotions}
\end{figure*}

To automatically change the illumination during broadcasting different emotions, a gesture recognition technique based on deep learning was implemented to run in parallel with body tracking (Fig. \ref{net}). 

\begin{figure}[H]
\centering
\includegraphics[width=1\linewidth]{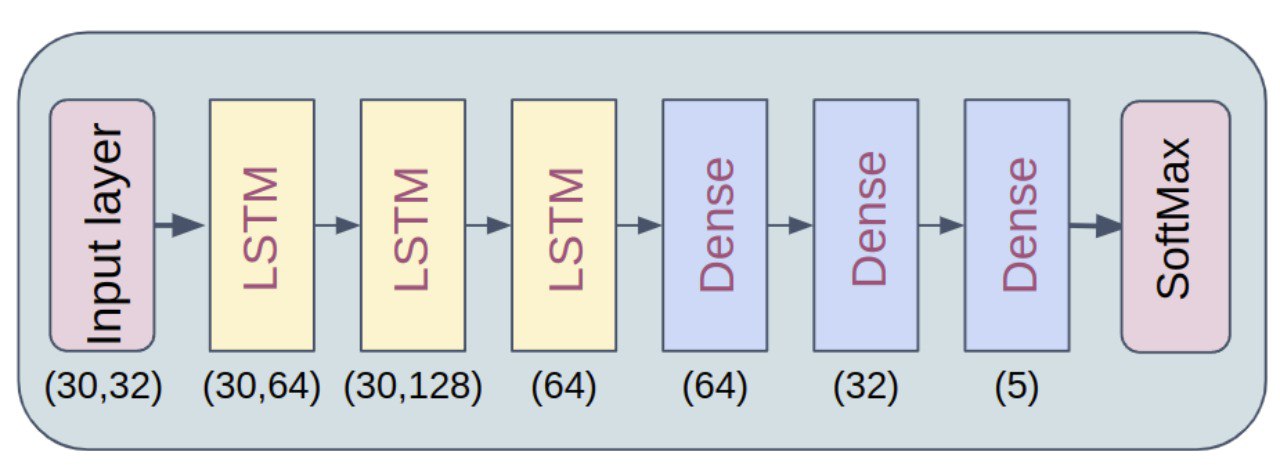}
\caption{Dynamic gesture recognition deep learning network architecture.} \label{net}
\end{figure}

The network was trained for 100 epochs (Fig. \ref{training}), achieving accuracy on the test dataset of 97\%. 
 
 \begin{figure}[H]
\centering
\includegraphics[width=0.8\linewidth]{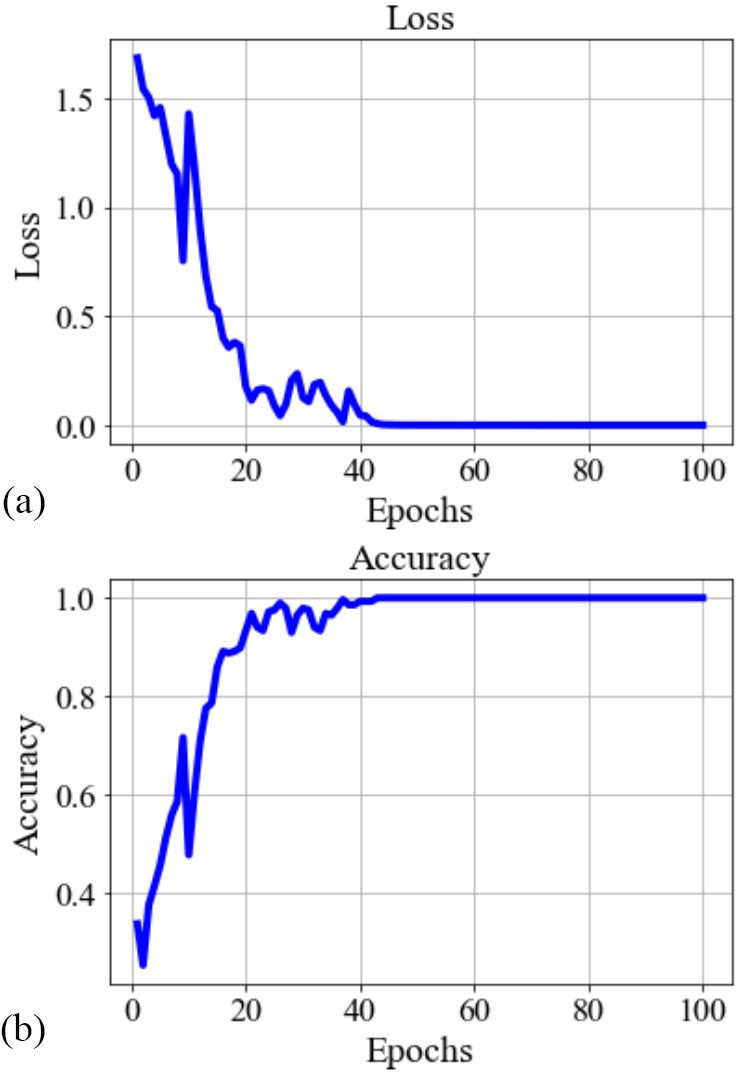}
\caption{Dynamic gesture recognition training. a) Loss. b) The categorical accuracy.} \label{training}
\end{figure}

The speed of drones within a swarm is limited. Therefore, the operator cannot change position instantly, which necessitates the use of dynamic gestures rather than static ones. The implemented gesture recognition technique utilizes a deep neural network with an architecture that uses Long short-term memory (LSTM) blocks to store data about the past positions of key points in a single sequence .

To train the neural network, sequences of body landmarks were used; each sequence consists of 30 video frames. Landmarks were extracted from each frame using the MediaPipe Holistic framework, as discussed in Section \ref{trajectory}. A custom dataset was collected from six participants. We gathered body landmarks from 600 collected videos for training on the recognition of our five basic gestures. 
 
Real-time recognition accuracy during user studies was 93\%. The user study was performed with a mixed group of participants; part of them didn't take part in recording the training dataset.

\section{Experimental Evaluation}
% Ahmed Baza, Aleksy Fedoseev

\emph{Participants}: We invited 10 participants (two females and eight males) aged from 22 to 26 (mean = 23.9, std = 1.22), to experience the visual interpretation of five different emotions from swarm avatar. Eight of the participants had not previously worked with drones, while two participants had interacted with drones several times. 
%They watched a series of poses and rated the degree to which they perceived most expressed emotion on a Likert scale (1 being strongly disagree, 3 being neutral and 5 being strongly agree).

 \emph{Procedure}: The experimental procedure applied for emotion evaluation is based on the methodology suggested by Ajili et al. \cite{Ajili_2018} for accessing virtual avatar expressive gestures performed with four basic emotions selected from Russell’s Circumplex model of affect (happy, sad, angry, and calm). In this paper, we proposed adding the fifth emotion of confusion to represent knowledge emotions \cite{Paul_2010} on top of the basic emotions mentioned above. Users watched a series of poses performed by the swarm avatar in simulation. %as shown in Fig. \ref{fig:parti}.

% \begin{figure}[!h]
% \centering
% \includegraphics[width=1\linewidth]{images/DSC_0996.JPG}
% \caption{User study setup.} \label{fig:parti}
% \end{figure}

Users were not being influenced by other external factors such as the sound or color of the drones. After watching the avatar performance, each user rated the degree to which they perceived expressed emotions on a 5-point Likert scale (1 being strongly disagreed, 3 being neutral and 5 being strongly agreed).

\emph{Experimental results}:

\begin{table}[!ht]
\centering
%\caption{Group percentage recognition of shape patterns}
\caption{Confusion Matrix of Emotion Recognition}
\label{tab:force}
%\resizebox{columnwidth}{!}{
%\scalebox{0.8}{
% \begin{tabular}{| c{50pt} | c{40pt} | c{40pt} | c{40pt} |}
\begin{tabular}{| c | c | c | c | c | c |}
\hline
\multicolumn{1}{|c|}{\cellcolor{Gray5}} &\multicolumn{5}{c|}{\textit{Estimated emotion}}\\
\hline
\textbf{}Performed emotion & Happy & Sad & Angry & Neutral & Confused \\
\hline
Happy & \cellcolor{Gray1}\textbf{4.50} & 1.20 &\cellcolor{Gray3}\textbf{2.80}& \cellcolor{Gray3}\textbf{2.10}&1.70 \\\hline
Sad & 1.90 & \cellcolor{Gray1}\textbf{4.20} &1.80& \cellcolor{Gray3}\textbf{2.70}& \cellcolor{Gray3}\textbf{2.40} \\\hline
Angry & \cellcolor{Gray3}\textbf{2.40} & \cellcolor{Gray3}\textbf{2.20} &\cellcolor{Gray1}\textbf{4.00}& 1.60& \cellcolor{Gray3}\textbf{2.90} \\\hline
Neutral & \cellcolor{Gray3}\textbf{2.80} & 1.80 &\cellcolor{Gray3}\textbf{2.40}& \cellcolor{Gray2}\textbf{3.70}&1.50 \\\hline
Confused & \cellcolor{Gray3}\textbf{2.50} & 1.80 &\cellcolor{Gray3}\textbf{2.40}& 1.50& \cellcolor{Gray2}\textbf{3.80} \\\hline

\end{tabular}
%}%}
\end{table}

To validate the internal consistency of user responses, we computed Cronbach’s alpha as a metric used to assess
the reliability of a set of Likert scale, it is expressed as a number between 0 and 1. The mean of Cronbach’s alpha is $> 0.8$ for all emotions, indicating the high consistency between the users' evaluation of the performed emotions. All emotions were evaluated relatively high by the users with the emotion of happiness being accessed higher than others (mean 4.5 out of 5).

\subsection{User evaluation of the SwarMan by NASA-TLX based Questionnaire}

All participants invited to the emotion recognition experiment were then invited to communicate with each other through the SwarMan interface. After the experiment, each participant was asked to complete a questionnaire based on The NASA Task Load Index (NASA-TLX) and three extra questions which give information such as age, gender of the participant, and previous experience with drones. An additional parameter of Intuitiveness was added to the questionnaire to evaluate how natural was avatar-based communication. Therefore, the participants provided feedback on seven questions: 
 
\begin{itemize}
  \item Mental Demand: How much mental and perceptual activity was required (e.g. deciding, calculating, etc)? Was the task easy or demanding, simple or complex? (Low - High)
    \item Physical Demand: How much physical activity was required? Was the task easy or demanding, slack or strenuous? (Low - High)
    \item Temporal Demand: How much time pressure did you feel due to the pace at which the tasks or task elements occurred? Was the pace slow or rapid? (Low - High)
    \item Overall Performance: How successful were you in performing the task? How satisfied were you with your performance? (Perfect - Failure)
    \item Effort: How hard did you have to work (mentally and physically) to accomplish your level of performance? (Low - High)
    \item Frustration Level: How irritated, stressed, and annoyed versus content, relaxed, and complacent did you feel during the task? (Low - High)
    \item Intuitiveness: How natural was the experience? How intuitive did you find the swarm avatar response to your emotions? (Intuitive - Artificial)
    \end{itemize}
    
The results of the NASA-TLX based survey are shown in Fig.~\ref{fig:likert}. %in Table~\ref{tab:table_NASA}. 

% \begin{table}[h]
%   \caption{Subjective feedback on 10-point NASA-TLX based Likert scale.}
%     \label{tab:table_NASA}
%     %\vspace{0.5 cm}
%     \begin{center}
%     \scalebox{0.9}{%
%     \begin{tabular}{|c||c||c||c|}
%     \hline
%       & Mean Score \\
%     \hline
%     Mental Demand & 2.25 (Low - High)\\
%     \hline
%     Physical Demand & 1.88  (Low - High)\\
%     \hline
%     Temporal Demand & 2.13  (Low - High\\
%     \hline
%     Overall performance & 1.38 ( Perfect - Failure)\\
%     \hline
%     Effort & 2.5  (Low - High)\\
%     \hline
%     Frustration & 1.4 (Low - High)\\
%         \hline
%     Intuitiveness & 1.5 (intuitive - Artificial)\\
%     \hline
%     \end{tabular}}
%     \end{center}
% \end{table}

\begin{figure}[htp]
\centering
\includegraphics[width=1\linewidth]{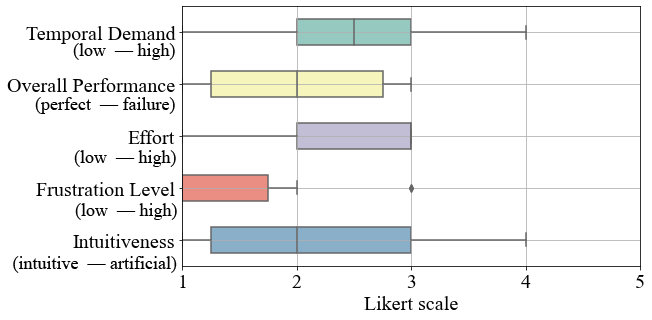}
\caption{Subjective feedback on the 5-point NASA-TLX based Likert scale.} \label{fig:likert}
\end{figure}

We conducted a chi-square analysis based on the frequency of answers in each category. The chi-square test of independence revealed that the participants' experience with drones does not affect the evaluation of swarm avatar control criteria, such as tiredness ($\tilde{\chi}^2$ = 2.66, $p$ = 0.92), temporal demand ($\tilde{\chi}^2$ = 2.3, $p$ = 0.94) and intuitiveness ($\tilde{\chi}^2$ = 1.33, $p$ = 0.98). In summary, participants did not feel any additional physical effort during the gesture control performance (mean of 1.9 on the 5-point scale). The majority of users did not experience Frustration (mean of 1.4 on the 5-point scale). In addition, the same users rated their overall performance satisfactory. All participants evaluated the system as intuitive (mean of 1.5 on the 5-point scale).

\section{Conclusion}%Table updated
We present a novel concept of telecommunications operating through an anthropomorphic drone swarm. The presented system utilized body tracking for trajectory generation to intuitively operate an anthropomorphic swarm of drones in a remote environment. Stable Control of the swarm of drones was realized using an artificial potential field algorithm and agent assignment based on minimum distance-based cost. The proposed system allows users to experience an effective human-swarm interaction utilizing the DNN-based gesture recognition technique. We achieved an acceptable gesture recognition accuracy of 93\% during our user studies, rendering the recognized gestures into illumination to enhance the user's affective experience. Based on the performed user studies, operators of the system have found it intuitive (1.5 on the Likert scale) with overall low physical and mental demand (1.88, 2.25 on the Likert scale respectively). 
With the high maneuverability, scalability, low cost, low weight, and small storing size, the usage of the SwarMan swarm-based avatar can potentially provide a solution for a wide scope of problems existing in current applications of anthropomorphic robots in telepresence. 

We plan to develop and integrate larger drones with higher payload in the system to allow operators physical interaction with the remote environment as well as with users through the sense of touch. The integration of more capable agents has the potential to increase the usability of the system and offers more solutions in telepresence applications.  
\section*{Acknowledgment}

The reported study was funded by RFBR and CNRS, project number 21-58-15006..

% \section*{References}

\bibliographystyle{IEEEtran}
\bibliography{bib}

% Generated by IEEEtran.bst, version: 1.14 (2015/08/26)
\begin{thebibliography}{10}
\providecommand{\url}[1]{#1}
\csname url@samestyle\endcsname
\providecommand{\newblock}{\relax}
\providecommand{\bibinfo}[2]{#2}
\providecommand{\BIBentrySTDinterwordspacing}{\spaceskip=0pt\relax}
\providecommand{\BIBentryALTinterwordstretchfactor}{4}
\providecommand{\BIBentryALTinterwordspacing}{\spaceskip=\fontdimen2\font plus
\BIBentryALTinterwordstretchfactor\fontdimen3\font minus
  \fontdimen4\font\relax}
\providecommand{\BIBforeignlanguage}[2]{{%
\expandafter\ifx\csname l@#1\endcsname\relax
\typeout{** WARNING: IEEEtran.bst: No hyphenation pattern has been}%
\typeout{** loaded for the language `#1'. Using the pattern for}%
\typeout{** the default language instead.}%
\else
\language=\csname l@#1\endcsname
\fi
#2}}
\providecommand{\BIBdecl}{\relax}
\BIBdecl

\bibitem{9501446}
M.~Abdullah, M.~Ahmad, and D.~Han, ``Hierarchical attention approach in
  multimodal emotion recognition for human robot interaction,'' in \emph{2021
  36th International Technical Conference on Circuits/Systems, Computers and
  Communications (ITC-CSCC)}, 2021, pp. 1--4.

\bibitem{6249581}
I.~Leite, G.~Castellano, A.~Pereira, C.~Martinho, and A.~Paiva, ``Modelling
  empathic behaviour in a robotic game companion for children: An ethnographic
  study in real-world settings,'' in \emph{2012 7th ACM/IEEE International
  Conference on Human-Robot Interaction (HRI)}, 2012, pp. 367--374.

\bibitem{8371114}
S.~Shimizu, K.~Shimada, and R.~Murakami, ``Non-verbal communication-based
  emotion incitation robot,'' in \emph{2018 IEEE 15th International Workshop on
  Advanced Motion Control (AMC)}, 2018, pp. 338--341.

\bibitem{hanson2020neurosymbolic}
D.~Hanson, A.~Imran, A.~Vellanki, and S.~Kanagaraj, ``A neuro-symbolic
  humanlike arm controller for sophia the robot,'' 2020.

\bibitem{lenz2021bimanual}
C.~Lenz and S.~Behnke, ``Bimanual telemanipulation with force and haptic
  feedback and predictive limit avoidance,'' in \emph{2021 European Conference
  on Mobile Robots (ECMR)}, 2021, pp. 1--7.

\bibitem{serpiva2021swarmpaint}
V.~Serpiva, E.~Karmanova, A.~Fedoseev, S.~Perminov, and D.~Tsetserukou,
  ``Swarmpaint: Human-swarm interaction for trajectory generation and formation
  control by dnn-based gesture interface,'' in \emph{2021 International
  Conference on Unmanned Aircraft Systems (ICUAS)}, 2021, pp. 1055--1062.

\bibitem{virtual_humans}
A.~Cordar, M.~Borish, A.~Foster, and B.~Lok, ``Building virtual humans with
  back stories: Training interpersonal communication skills in medical
  students,'' in \emph{Intelligent Virtual Agents}, T.~Bickmore, S.~Marsella,
  and C.~Sidner, Eds.\hskip 1em plus 0.5em minus 0.4em\relax Cham: Springer
  International Publishing, 2014, pp. 144--153.

\bibitem{Tachi_2020}
\BIBentryALTinterwordspacing
S.~Tachi, Y.~Inoue, and F.~Kato, ``Telesar vi: Telexistence surrogate
  anthropomorphic robot vi,'' \emph{International Journal of Humanoid
  Robotics}, vol.~17, no.~05, p. 2050019, 2020. [Online]. Available:
  \url{https://doi.org/10.1142/S021984362050019X}
\BIBentrySTDinterwordspacing

\bibitem{Tsetserukou_2010}
D.~Tsetserukou and A.~Neviarouskaya, ``ifeel\_im!: Augmenting emotions during
  online communication,'' \emph{IEEE Computer Graphics and Applications},
  vol.~30, no.~5, pp. 72--80, 2010.

\bibitem{Bartneck_2004}
C.~Bartneck, J.~Reichenbach, and A.~van Breemen, ``In your face, robot! the
  influence of a character’s embodiment on how users perceive its emotional
  expressions.'' 2004.

\bibitem{Chao-gang_2008}
W.~Chao-gang, Z.~Jie-yu, and Z.~Yuan-yuan, ``An emotion generation model for
  interactive virtual robots,'' in \emph{2008 International Symposium on
  Computational Intelligence and Design}, vol.~2, 2008, pp. 238--241.

\bibitem{Vasquez_2020}
\BIBentryALTinterwordspacing
B.~P.~E. {Alvarado Vásquez} and F.~Matía, ``A tour-guide robot: Moving
  towards interaction with humans,'' \emph{Engineering Applications of
  Artificial Intelligence}, vol.~88, p. 103356, 2020. [Online]. Available:
  \url{https://www.sciencedirect.com/science/article/pii/S0952197619302908}
\BIBentrySTDinterwordspacing

\bibitem{Matsui_2005}
D.~Matsui, T.~Minato, K.~MacDorman, and H.~Ishiguro, ``Generating natural
  motion in an android by mapping human motion,'' in \emph{2005 IEEE/RSJ
  International Conference on Intelligent Robots and Systems}, 2005, pp.
  3301--3308.

\bibitem{Cohen_2011}
I.~Cohen, R.~Looije, and M.~A. Neerincx, ``Child's recognition of emotions in
  robot's face and body,'' in \emph{2011 6th ACM/IEEE International Conference
  on Human-Robot Interaction (HRI)}, 2011, pp. 123--124.

\bibitem{Yoon_2019}
Y.~Yoon, W.-R. Ko, M.~Jang, J.~Lee, J.~Kim, and G.~Lee, ``Robots learn social
  skills: End-to-end learning of co-speech gesture generation for humanoid
  robots,'' in \emph{2019 International Conference on Robotics and Automation
  (ICRA)}, 2019, pp. 4303--4309.

\bibitem{Marmpena_2019}
M.~Marmpena, A.~Lim, T.~S. Dahl, and N.~Hemion, ``Generating robotic emotional
  body language with variational autoencoders,'' in \emph{2019 8th
  International Conference on Affective Computing and Intelligent Interaction
  (ACII)}, 2019, pp. 545--551.

\bibitem{Ajili_2018}
I.~Ajili, M.~Mallem, and J.-Y. Didier, ``Human motions and emotions recognition
  inspired by lma qualities,'' \emph{The Visual Computer}, pp. 1--16, 2018.

\bibitem{Paul_2010}
P.~Silvia, ``Confusion and interest: The role of knowledge emotions in
  aesthetic experience,'' \emph{Psychology of Aesthetics, Creativity, and the
  Arts}, vol.~4, pp. 75--80, 05 2010.

\end{thebibliography}

\end{document}